\documentclass[
  manuscript=preprint,  
]{cup-journal}

\pdfoutput=1           
\usepackage{amsmath}
\usepackage[nopatch]{microtype}
\usepackage{booktabs}
\usepackage{hyperref}
\usepackage{listings}
\usepackage{xcolor}
\usepackage{amssymb}

\title{Building a Strong Instruction Language Model for a Less-Resourced Language}

\author{Domen Vreš}
\affiliation{University of Ljubljana, Faculty of Computer and Information Science, Ljubljana, Slovenia}
\email[D. Vreš]{domen.vres@fri.uni-lj.si}

\author{Tjaša Arčon}
\affiliation{University of Ljubljana, Faculty of Computer and Information Science, Ljubljana, Slovenia}

\author{Timotej Petrič}
\affiliation{University of Ljubljana, Faculty of Computer and Information Science, Ljubljana, Slovenia}

\author{Dario Vajda}
\affiliation{University of Ljubljana, Faculty of Computer and Information Science, Ljubljana, Slovenia}

\author{Marko Robnik-Šikonja}
\affiliation{University of Ljubljana, Faculty of Computer and Information Science, Ljubljana, Slovenia}

\author{Iztok Lebar Bajec}
\affiliation{University of Ljubljana, Faculty of Computer and Information Science, Ljubljana, Slovenia}


\begin{document}

\begin{abstract}
Large language models (LLMs) have become an essential tool for natural language processing and artificial intelligence in general. Current open-source models are primarily trained on English texts, resulting in poorer performance on less-resourced languages and cultures. We present a set of methodological approaches necessary for the successful adaptation of an LLM to a less-resourced language, and demonstrate them using the Slovene language. We present GaMS3-12B, a generative model for Slovene with 12 billion parameters, and demonstrate that it is the best-performing open-source model for Slovene within its parameter range. We adapted the model to the Slovene language using three-stage continual pre-training of the Gemma 3 model, followed by two-stage supervised fine-tuning (SFT). We trained the model on a combination of 140B Slovene, English, Bosnian, Serbian, and Croatian pretraining tokens, and over 200 thousand English and Slovene SFT examples. We evaluate GaMS3-12B on the Slovenian-LLM-Eval datasets, English-to-Slovene translation, and the Slovene LLM arena. We show that the described model outperforms 12B Gemma 3 across all three scenarios and performs comparably to much larger commercial GPT-4o in the Slovene LLM arena, achieving a win rate of over 60 \%.
\end{abstract}

\section{Introduction}
\label{sec:intro}

Large language models (LLMs), particularly generative LLMs like GPT models, have dramatically transformed natural language processing (NLP), advancing our understanding and ability to generate human language. As a result of this rapid development, new open-source decoder-type transformer LLMs such as Gemma~\citep{gemma}, Llama~\citep{llama}, Nemotron~\citep{nemotron}, and many others are released on a regular basis. These models are trained predominantly on high-resource languages (primarily English) while their attention and performance on less-resourced languages, such as Slovene, is limited. To amend this situation, we present the development of \textbf{GaMS3-12B} (\textbf{G}ener\textbf{a}tive \textbf{M}odel for \textbf{S}lovene), an open-source generative model for Slovene with 12 billion parameters. We evaluate GaMS3-12B on a suite of Slovene benchmarks, Slovenian-LLM-Eval~\citep{slo_eval}, English to Slovene translation, and using the Slovene LLM Arena\footnote{\url{https://arena.cjvt.si/en/leaderboard}}. We demonstrate clear improvement over the base Gemma 3 model, providing a strong indication that it is possible to enhance LLMs for many other less-resourced languages. As many such developments are ongoing or being planned, the description of our development has considerable generalization value beyond a single language.

The primary challenge in training LLMs for less-resourced languages is the scarcity of data. For example, the Llama 4 model was trained on 30 trillion tokens, whereas the currently available authentic Slovene corpora contain around 40 billion tokens or approximately one thousand times less. This means that training an LLM from scratch for Slovene (and other less-resourced languages) is unfeasible. Hence, we take the pretrained Gemma 3 model and adapt it for Slovene. We perform three continual pretraining (CPT) stages, followed by two supervised fine-tuning (SFT) stages. To increase the amount of training data, we also include texts from similar languages, i.e., Croatian, Bosnian, and Serbian, which can improve the models' performance. Additionally, we generate more data by performing an OCR of high-quality Slovene PDFs, such as the repositories of top Slovene universities and the digital archive of the national library. We also translate high-quality English corpora such as Wikipedia and Nemotron datasets.

Our contributions are as follows:
\begin{itemize}
    \item GaMS, a State-of-the-art open-source generative model for Slovene;
    \item An open-source code for the OCR of Slovene documents, that can be transferred to other less-resourced languages;
    \item A clear description of data generation and preparation methodology, together with model training steps, making our work transferable to other less-resourced languages.
\end{itemize}

Developing open-source models for less-resourced languages is crucial for preserving the autonomy, cultural identity, and sovereignty of such languages. As these languages are currently only sufficiently supported in commercial models, users who want to use them are forced to send their data to large companies. In contrast, for languages where open LLMs exist, companies and public institutions can set up open-source models on-premises, retaining full control over their customers' data residency. Additionally, the creation of the Slovene GaMS model is open, well-documented, and offers useful lessons to other less-resourced languages.

The rest of the paper is organized as follows. In Section \ref{sec:related_work}, we present relevant related work for language adaptation of large language models. We offer a detailed description of our continual pretraining approach and data in Section \ref{sec:cpt}. We describe supervised fine-tuning data and approach in Section \ref{sec:sft}. In Section \ref{sec:training}, we present the training environment, hardware resources, and hyperparameters. We evaluate our model in Section \ref{sec:evaluation}, and offer conclusions and ideas for future work in Section \ref{sec:conclusion}.

\section{Related work}
\label{sec:related_work}

Most state--of--the--art language models~\citep{gpt3,gemini,llama,gemma} are trained using the same pipeline. The models first go through the pretraining phase, where they are trained to predict the next token on huge corpora. This is followed by the supervised fine-tuning (SFT) phase, where models are trained to respond to users' prompts. Finally, models are trained to produce outputs that align with user preferences (e.g., safer, factually correct, and helpful responses) using methods such as reinforcement learning with human feedback (RLHF). We follow a similar pipeline, but replace the reinforcement learning with chat-enhancing SFT due to a lack of preference data for Slovene.

Multilingual models were developed to extend the benefits of LLMs to languages that lack sufficient training data. These models are typically based on English with additional training data in other languages. The best performing open-source multi-lingual models for Slovene are EuroLLM~\citep{euro_llm}, Gemma 2~\cite{gemma}, Gemma 3~\citep{gemma3}, DeepSeek~\citep{deepseek}, Qwen~\citep{qwen3-235b-a22b-yang2025qwen3}, Llama 4~\citep{llama4}, Mistral~\citep{mistral}, and Apertus~\citep{apertus}. However, the models such as DeepSeek, Llama 4, and the largest versions of Mistral and Qwen are often infeasible to run on-premises due to a large number of parameters that require a significant amount of GPU vRAM. Smaller versions of Mistral and Qwen, on the other hand, perform significantly worse on Slovene and are beaten by other models of similar size. For Slovene, the most interesting family of models mentioned above is Gemma due to its strong performance and EuroLLM due to its focus on EU languages. EuroLLM was trained from scratch on a data mixture consisting of $50~\%$ of English texts, $5~\%$ of code/math data, and $45~\%$ of other languages. These other languages represent the majority of European languages. Slovene data represents $1~\%$ of the other languages' data. The percentage of Slovene data used for pretraining of Gemma models is unclear, but it is significant enough that the 27B version of Gemma 3 performs better than the rest of multi--lingual models in the Slovenian-LLM-Eval~\citep{slo_eval} suite of benchmarks (see Table \ref{tab:slo_eval_leaderboard}). This justifies our choice of Gemma 3 as our starting model. However, the Gemma 3 models still benefit from training on more Slovene data, as clearly shown by the gain in GaMS3 performance compared to Gemma 3 12B.

Over the last few years, some decoder-type English models have been adapted for specific languages. The first such models were GPT-SW3 \citep{gpt-sw3} for Swedish, Chinese LLaMa \citep{chinese_llama} and Open-Chinese-LLaMA \citep{open_chinese_llama} for Chinese, and Gervasio \citep{gervasio} for Portuguese. However, these models were either trained from scratch (GPT-SW3), performed vocabulary expansion without embedding transfer, resulting in knowledge loss (Chinese LLaMA and Open-Chinese-LLaMA), or were instruction-tuned (SFT) for the target language (Gervasio). In contrast, we first continually pretrained the Gemma 3 model, which we believe is essential to adapt the model to Slovene culture, and then performed SFT. Recently, NVIDIA started the Sovereign AI initiative, supporting model builders for sovereign languages across Europe, the Middle East, and Asia. Apart from the GaMS family, the most notable representatives of this initiative are Falcon~\citep{falcon} and Bielik~\citep{bielik_v3}. While Falcon is trained from scratch and focuses on multi--linguality, the Bielik team used a similar approach to ours. They started with a pretrained Mistral model and adapted it for Polish through CPT, SFT, and reinforcement learning. While Bielik shows decent performance on Slovene, it does not achieve the same level of performance as Slovene specialized models.

Several LLMs were already developed for Slovene. Encoder-type and encoder-decoder-type models include CroSloEngual BERT~\citep{cse_bert}, SloBERTa~\citep{sloberta}, and SloT5~\citep{slot5}. The first decoder-type models adapted for Slovene were SlovenianGPT~\citep{slovenian_gpt}, GaMS-1B~\citep{GaMS}, and TinySLlama~\citep{tiny_sllama}. Due to their older architecture, shorter contexts, and limited training data, these models are less useful in practice compared to current models. However, they provide valuable insights into the Slovenian specialization. They show the benefits of training the model combined with languages similar to Slovene, such as Croatian (CroSloEngual BERT), and the impact of vocabulary modification and embedding transfer (GaMS-1B). Currently, the most relevant Slovene specialized decoder-type LLMs are Gemma 2 based GaMS models~\citep{gams_9b} with 9B and 27B parameters and Mistral NeMo based Zlatorog 12B~\citep{zlatorog}. The best performing model out of these is GaMS-27B-Instruct-Nemotron~\citep{gams_27b_nemotron}. Compared to GaMS3, all of these models were trained on much less data (around 15B tokens) and on much shorter contexts (Zlatorog has the longest context window with 16k tokens), making them useless on longer texts. GaMS3, on the other hand was trained on around 140B tokens and supports the context window of 131k tokens.

\section{Continual pretraining}
\label{sec:cpt}
The adaptation of an LLM to a specific language can be split into two phases: continual pretraining and supervised fine-tuning. We describe the first phase in this section, while the second phase is presented in Section \ref{sec:sft}.

The continual pretraining (CPT) stage is necessary to improve a model's understanding of the target language and culture. Our previous work on GaMS models~\citep{GaMS,gams_9b} showed the improvement in Slovene benchmark performance after CPT. With GaMS3 we split CPT into three stages: parallel alignment, base CPT, and long CPT. Each stage uses a different combination of data and serves a specific purpose in enhancing the base model’s understanding of the Slovene language and culture. We describe the parallel alignment stage in Section \ref{sec:parallel_alignment}, the base CPT stage in Section \ref{sec:base_cpt}, and the long CPT stage in Section \ref{sec:long_cpt}. To obtain more CPT data, we performed OCR of high-quality Slovene PDFs. We describe our OCR pipeline in Section \ref{sec:slovene_ocr}. To enhance the training process and make it more efficient, we employed various data preparation techniques, which we describe in Section \ref{sec:cpt_data_prep}.

\subsection{Parallel alignment}
\label{sec:parallel_alignment}
The purpose of the parallel alignment stage is to explicitly align the English and Slovene languages inside the model. This stage was inspired by \citet{align_after_pretrain}, who performed language alignment after the pretraining phase by combining multilingual contrastive learning and cross-lingual instruction tuning. The former tries to increase the similarity between the model's representation of the same sentence in the source and target languages. The latter provides the model with a prompt in the source language and the response in the target language. We moved the parallel alignment to the first CPT stage to bridge the gap between the Gemma 3 training distribution (which was dominated by the English language) and our training distribution (dominated by the Slovene language). This approach has already shown promising results when training the previous generation of GaMS models (GaMS-9B and GaMS-27B).

To perform parallel alignment, we used a combination of the following parallel English and Slovene corpora:
\begin{itemize}
    \item \textbf{Human translated corpora}: KAS Abstracts~\citep{kas_abs}, DGT~\citep{dgt} and MaCoCu~\citep{macocu_par};
    \item \textbf{Machine translated corpus}: Wikipedia
\end{itemize}

Here, we converted Wikipedia into a parallel corpus by machine translating English Wikipedia in markdown format~\citep{wikipedia_markdown} to Slovene using our GaMS-9B-SFT-DPO-Translator~\citep{gams_translator}. The tokenized data sizes for the parallel corpora are shown in Table \ref{tab:parallel_stats}. We trained the model with the context window of 65k tokens during this stage.

\begin{table}[hbt!]
\begin{threeparttable}
\caption{Number of tokens and training documents for parallel alignment corpora. The numbers are displayed after tokenization using the Gemma 3 tokenizer, sequence packing, and padding.}
\label{tab:parallel_stats}
\begin{tabular}{lrrr}
\toprule
\headrow Corpus & Number of tokens & Number of documents & Total percentage \\
\midrule
DGT & 804,847,616 & 12,281 & 6.3 \% \\
MaCoCu & 430,374,912 & 6,567 & 3.4 \% \\
KAS & 31,391,744 & 479 & 0.2 \%  \\
Wikipedia & 11,529,093,120 & 175,920 & 90.1 \% \\
\midrule
\textbf{Total} & \textbf{12,795,707,392} & \textbf{195,247} & \\
\bottomrule
\end{tabular}
\end{threeparttable}
\end{table}

We aligned English and Slovene texts on three levels. In the paragraph-level alignment, paragraphs of parallel documents were concatenated. A paragraph in one language was followed by the same paragraph in the other language, followed by the next paragraph in the first language, etc. In the document-level alignment, two documents in different languages were concatenated into a single training example. In separate documents, documents in different languages were used as different training examples. The alignment levels were chosen based on the dataset format and average document length.

\subsection{Base CPT}
\label{sec:base_cpt}
The idea behind the base CPT stage is to first perform a CPT with a shorter context to increase the efficiency of the training. We used a context window of 65k tokens during this stage. We chose such a context window so that the training still fitted on the available GPU hardware (i.e., LEONARDO Booster partition). We opted not to use a shorter context window at this stage, as our pretraining mix included longer documents. Moreover, Gemma 3 supports a context window of 131k tokens and we did not want to override its long context capabilities by using too short sequence length. During this stage, the model was trained on general corpora of Slovene, English, Croatian, Bosnian, and Serbian. We included English corpora to prevent forgetting, while Croatian, Bosnian, and Serbian corpora were utilized due to their cultural and language similarity to Slovene and additional support for code-mixed texts. Previous research~\citep{cse_bert,GaMS} shows that adding such languages is beneficial for the model's performance in Slovene, and it also leads to decent performance in these similar languages.

We combined the following types of data in the base CPT mix:
\begin{itemize}
    \item \textbf{NVIDIA's Nemotron pretraining datasets}~\citep{nemotron_nano2, nemotron_cc}: combination of English pretraining datasets released by NVIDIA. We selected a subset of splits from the released dataset and randomly subsampled data from each split to ensure that the majority of the CPT data is Slovene.
    \item \textbf{Collection of Slovene corpora from previous projects}: metaFida~\citep{metafida}.
    \item \textbf{Academic data}: KAS~\citep{kas}, documents from the repositories of Slovene universities (mostly BSc, BA, MSc, MA, and PhD theses), with the cutoff year of 2018.
    \item \textbf{News data}: Trendi~\citep{trendi}, the cutoff date at this stage is December 2023 (newer data was used in the long CPT stage).
    \item \textbf{Legal data}: Combination of Slovene legal data from public sources such as PISRS~\footnote{\url{https://pisrs.si}} and Uradni list~\footnote{\url{https://www.uradni-list.si}}.
    \item  \textbf{Medical data}: Combination of Slovene medical books, blogs, journals, papers, and articles.
    \item \textbf{National library data}: Data from the digital archive of the Slovene National and University Library (NUK). The used data is not copyright-protected and consists of old newspapers, journals, and books. There are also some modern scientific journals or papers.
    \item \textbf{Wikipedia}: Slovene, Croatian, Bosnian, and Serbian Wikipedia dumps in markdown format. The dump date for all languages is January 1st, 2025.
    \item \textbf{Web crawls}: Slovene parts of CLASSLA-web~\citep{classla_web} and FineWeb2~\citep{fineweb2}. We enhanced a part of the CLASSLA corpus (1 million documents) using Gemma 3 27B and a data generation prompt based on the ReWire method~\citep{rewire};
    \item \textbf{Web PDFs}: Slovene, Croatian, Bosnian, and Serbian splits of the FinePDFs corpus~\citep{finepdfs}. Approximately 75\% of the data is used at this stage, while the remaining 25\% is saved for the long CPT stage.
\end{itemize}

The tokenized data sizes for the base CPT corpora are shown in Table \ref{tab:base_cpt_stats}.

\begin{table}[hbt!]
\begin{threeparttable}
\caption{Number of tokens and training documents for the base CPT corpora. The numbers are shown after the tokenization using the Gemma 3 tokenizer, sequence packing and padding. Wikipedia-Yugo stands for combined Slovenian, Bosnian, Croatian and Serbian Wikipedia dums. NUK stands for the Slovenian National and University Library.}
\label{tab:base_cpt_stats}
\resizebox{\textwidth}{!}{
\begin{tabular}{l l r r r}
\hline
\headrow Corpus & Language & Number of tokens & Number of documents & Total percentage \\
\hline
Nemotron-Pretraining-Code & English & 1,952,120,832 & 29,787 & 1.9 \% \\
Nemotron-CC-Math-v1-4Plus & English & 2,526,937,088 & 38,558 & 2.5 \% \\
Nemotron-CC-Math-v1-3 & English & 1,210,908,672 & 18,477 & 1.2 \% \\
Nemotron-Pretraining-SFT-v1 & English & 3,718,316,032 & 56,737 & 3.7 \% \\
Nemotron-CC-v2-HighQuality-Synthetic & English & 10,479,403,008 & 159,903 & 10.4 \% \\
Nemotron-CC-v2-DiverseQA & English & 8,631,353,344 & 131,704 & 8.6 \% \\
FinePDFs-Bos & Bosnian & 4,815,912,960 & 73,485 & 4.8 \% \\
FinePDFs-Hrv & Croatian & 9,541,124,096 & 145,586 & 9.5 \% \\
FinePDFs-Srp & Serbian & 8,119,844,864 & 123,899 & 8.0 \% \\
FinePDFs-Slv & Slovenian & 5,925,044,224 & 90,409 & 5.9 \% \\
Trendi & Slovenian & 1,737,687,040 & 26,515 & 1.7 \% \\
CLASSLA-Web-2024 & Slovenian & 4,256,432,128 & 64,948 & 4.2 \% \\
Sl-Legal & Slovenian & 1,697,710,080 & 25,905 & 1.7 \% \\
Sl-Med & Slovenian & 1,598,095,360 & 24,385 & 1.6 \% \\
metaFida & Slovenian & 4,591,910,912 & 70,067 & 4.6 \% \\
FineWeb2 & Slovenian & 13,890,289,664 & 211,949 & 13.8 \% \\
KAS & Slovenian & 2,726,035,456 & 41,596 & 2.7 \% \\
NUK & Slovenian & 12,784,041,984 & 195,069 & 12.7 \% \\
Wikipedia-Yugo & Slovenian, Croatian, Bosnian, Serbian & 673,775,616 & 10,281 & 0.7 \% \\
\hline
\textbf{Total} &  & \textbf{100,876,943,360} & \textbf{1,539,260} &  \\
\hline
\end{tabular}}
\end{threeparttable}
\end{table}

\subsection{Long CPT}
\label{sec:long_cpt}

In the final, long CPT stage, the context window was increased to 131k tokens to capture the full Gemma 3 context window. In this stage, only the following, highest-quality, corpora were used:
\begin{itemize}
    \item \textbf{Nemotron corpora}: compared to the base CPT stage, we omitted the Math-3 and Pretraining-Code splits. For the remaining splits, we used smaller and disjoint subsets than in the base CPT stage. Additionally, we translated the Math-4 and Pretraining-SFT splits into Slovene using our GaMS-9B-SFT-DPO-Translator. We translated the former to increase the amount of Slovene math data and the latter to add some pretaining SFT data in Slovene, which can be beneficial.
    \item \textbf{FinePDFs}: we used the remaining 25\% of Slovene, Croatian, Bosnian and Serbian FinePDFs data.
    \item \textbf{News data}: we used the news articles from the Trendi corpus between January 2024 and July 2025.
    \item \textbf{Academic data}: since the KAS corpus covers academic works only up to the year 2018, we decided to OCR the PDFs collected from the repositories of Slovene universities. We included works published between 2019 and 2024. We name this corpus KAS-Extension.
    \item \textbf{Slovene math data}: we OCRed two Slovene Math and Physics journals Presek~\footnote{\url{http://www.presek.si/predstavitev.php}} and OMF~\footnote{\url{http://www.obzornik.si}}. Since these two journals combined give fewer than 100 million words, we used translations of Nemotron Math to construct the Slovene math corpus.
\end{itemize}

The tokenized data sizes for the long CPT corpora are shown in Table \ref{tab:long_cpt_stats}.

\begin{table}[hbt!]
\begin{threeparttable}
\caption{Number of tokens and training documents for the long CPT corpora. The numbers are shown after the tokenization using the Gemma 3 tokenizer, sequence packing, and padding.}
\label{tab:long_cpt_stats}
\resizebox{\textwidth}{!}{
\begin{tabular}{l l r r r}
\hline
\headrow Corpus & Language & Number of tokens & Number of documents & Total percentage \\
\hline
Nemotron-CC-Math-v1-4Plus & English & 1,087,373,312 & 8,296 & 5.4 \% \\
Nemotron-Pretraining-SFT-v1 & English & 1,231,945,728 & 9,399 & 6.1 \% \\
Nemotron-CC-v2-HighQuality-Synthetic & English & 2,634,285,056 & 20,098 & 13.1 \% \\
Nemotron-CC-v2-DiverseQA & English & 1,237,975,040 & 9,445 & 6.2 \% \\
FinePDFs-Bos & Bosnian & 1,614,282,752 & 12,316 & 8.0 \% \\
FinePDFs-Hrv & Croatian & 2,385,248,256 & 18,198 & 11.9 \% \\
FinePDFs-Srp & Serbian & 2,074,345,472 & 15,826 & 10.3 \% \\
FinePDFs-Slv & Slovenian & 1,969,618,944 & 15,027 & 9.8 \% \\
Trendi & Slovenian & 610,533,376 & 4,658 & 3.0 \% \\
KAS-Extension & Slovenian & 2,256,404,480 & 17,215 & 11.2 \% \\
Sl-Math & Slovenian & 1,456,078,848 & 11,109 & 7.2 \% \\
Nemotron-Pretraining-SFT-Translation & Slovenian & 1,553,858,560 & 11,855 & 7.7 \% \\
\hline
\textbf{Total} &  & \textbf{20,111,949,824} & \textbf{1,539,260} &  \\
\hline
\end{tabular}}
\end{threeparttable}
\end{table}

\subsection{OCR of Slovene PDFs}
\label{sec:slovene_ocr}

We performed OCR on the PDFs from three different sources:
\begin{itemize}
    \item \textbf{National library data}: We obtained a large amount of old newspapers, books, journals, articles, and other materials from the national library in PDF format, which in most cases represented scans of the original works. Some of the obtained PDF documents were accompanied by the OCR versions in txt format (created with the ABBYY tool). Since these were quite old, with low overall quality, poor formatting, and numerous artifacts (especially in the case of books), we decided to repeat the OCR with modern OCR tools. Among all our collections, this one is the largest and the most challenging for OCR.
    \item \textbf{Final theses from Slovene universities}: We collected PDFs published between 2019 and 2024 from the repositories of the top 3 Slovene universities. Even though this dataset is also large, it is less challenging than the national library one because the PDFs were digitally produced and their contents have a more structured layout.
    \item \textbf{Slovene math and physics journals}: We collected the PDFs of two Slovene math and physics journals: Presek and Obzornik za matematiko in fiziko. They have a slightly more challenging layout than university theses, but the PDFs were digitally produced, and the dataset size is small.
\end{itemize}

\begin{figure}[hbt!]
\centering
\includegraphics[width=\linewidth]{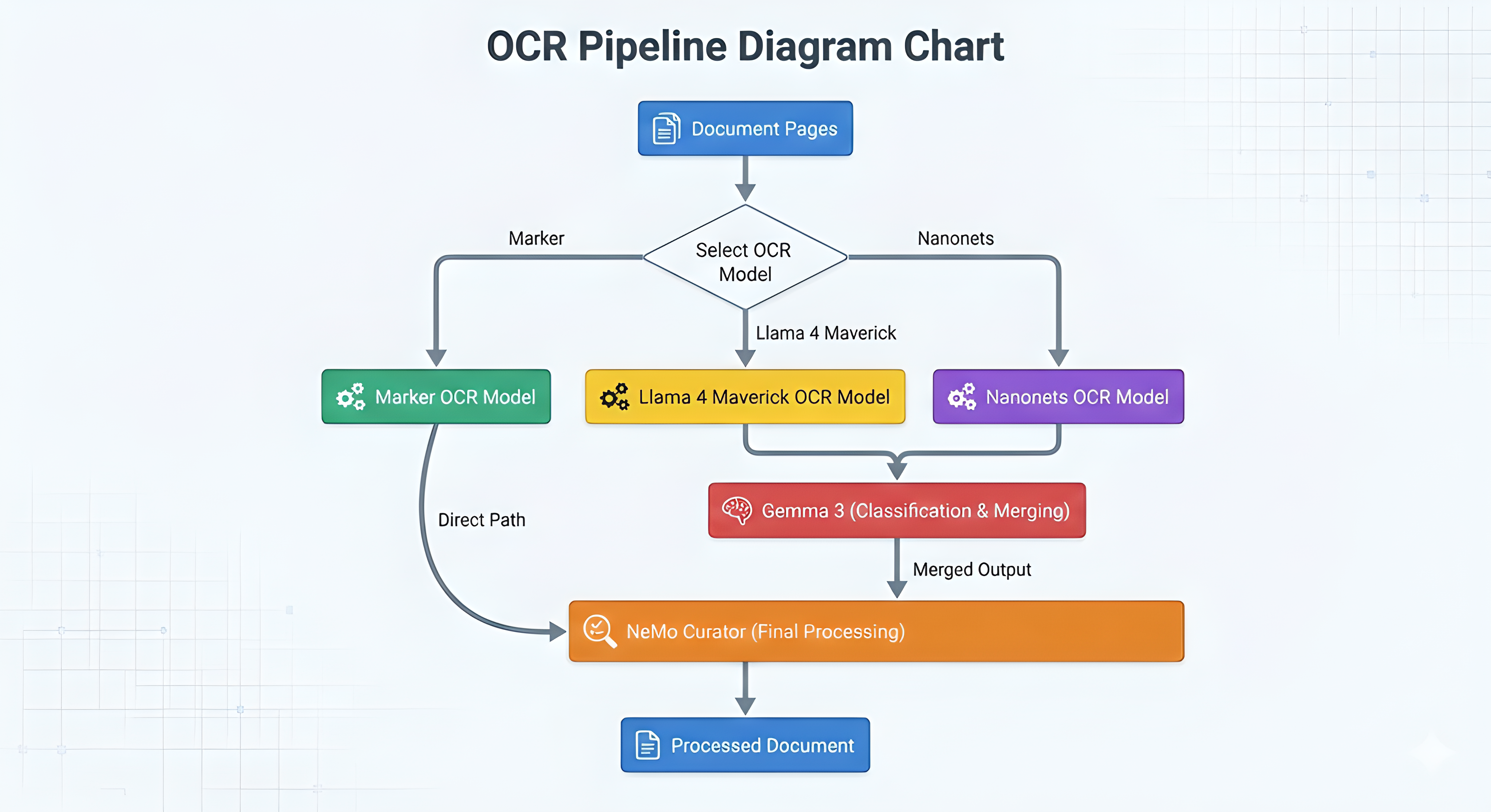}
\caption{Our three-stage OCR pipeline. Each PDF document is first converted to markdown using an OCR model. In case of Llama 4 and Nanonets, the document is OCRed page by page, and the pages are then post-processed using Gemma 3 27B it. With the marker library, these two stages are merged inside the library. In the final stage, the text cleaning is performed using NeMo Curator filters. The diagram chart was AI-generated using the Nano Banana Pro tool.}
\label{fig:ocr_cbart}
\end{figure}

We used a three-stage OCR pipeline and we are releasing the source code under an open-source license~\footnote{\url{https://github.com/GaMS-Team/local_ocr}}. In the first stage, the PDFs are converted to markdown using an OCR model. In the second stage, the markdown is post-processed using a separate language model. In the third stage, the markdowns are cleaned using the NeMo Curator tool. We present our pipeline as a diagram chart in Figure \ref{fig:ocr_cbart} and describe it in the three following subsections.

\subsubsection{OCR stage}

We preliminarily tested different OCR models and, using manual inspections and an internal benchmarks, we selected three that are useful for our task:
\begin{itemize}
    \item \textbf{Marker~\footnote{\url{https://github.com/datalab-to/marker}}} is a PDF to markdown conversion library that converts the whole document (hence one can avoid per-page post-processing, as it is already handled by the marker library). Besides the final markdown, the library also extracts the figures and images in PNG format and includes them in the markdown. The library internally runs several different vision models and offers an LLM postprocessing either through a commercial API or a local server. We skipped the LLM option here, as our testing showed that it did not significantly improve the results, but it did significantly slow down the conversion. We used the library at version \texttt{1.6.2}. Out of the three used OCR methods, this one is the fastest to run, but the most sensitive to different layouts and difficult formats.
    \item \textbf{Llama-4-Maverick}~\citep{llama4} is a multi-modal and multi-lingual LLM released by Meta, that shows good OCR performance. We converted the PDFs into markdown by sending documents page-by-page to the model. Each page is sent as a base64 encoded image, and the model is instructed to extract the text in markdown format using the prompt from \ref{app:llama_ocr}. Llama 4 shows superior performance to the other two models on academic works, where the layout structure is well-defined. As a plus, it is also capable of accurately extracting equations and tables from PDFs. However, it performs worse than Nanonets with difficult layouts (especially old newspapers, where text is all over the place). It is also the most computationally demanding and the slowest one to run. It requires a system with at least 640GB of combined vRAM (e.g. an entire node with 8xH100 GPUs with 80GB each, or four B200 GPUs with 180 GB of vRAM each). We set it up on a DGX B200 node on our internal GPU cluster.
    \item \textbf{Nanonets-OCR-s}~\citep{nanonets} is a fine-tuned version of Qwen2.5-VL-3B-Instruct multi-modal model. The model was fine-tuned on a curated dataset consisting of research papers, financial documents, legal documents, healthcare documents, tax forms, receipts, and invoices. It supports equation transcription (LaTeX), image captioning, watermark extraction, and the recognition of complex tables. Similarly to Llama 4, we sent each PDF document page-by-page to the model and encoded as a base64 image. We used the default prompt, as suggested in the models' card (see \ref{app:nanonets_ocr} for the exact prompt). Compared to Llama 4 this model is faster and less comuptaationally demanding to run. It handles difficult layouts well, but it tends to insert artifacts into the text and can start cycling due to its size. Luckily, such artifacts and cycles are easy to detect, and they can be removed during the filtering stage.
\end{itemize}

Note that we initiated our OCR process in May 2025, when many current OCR models, such as DeepSeekOCR \citep{deepseek_ocr}, had not yet been released. Moreover, commercial models such as GPT and Gemini might perform even better, but their usage is not feasible due to the data size and privacy issues.

We used Llama 4 for the OCR of academic works and the math and physics journals. For the National Library data, we used the following approach. For works that have the title of the newspaper/journal/book annotated (we did not receive any metadata from the National Library), we first performed OCR using the marker library. We then inspected a few examples for each newspaper/journal/book, and if they were of poor quality, repeated the process using either Llama 4 (for simpler layouts) or Nanonets (for the most challenging layouts). We sent the PDFs without any metadata directly through Nanonets, as it offers the highest quality to speed ratio among the three models.

\subsubsection{Post-processing stage}
Since the PDFs were sent through the Llama 4 and Nanonets as separate pages, these pages need to be merged back into a single document in the post-processing stage using Gemma 3 27B it model. We choose Gemma 3 as it is small enough to process a large amount of data in a reasonable time, it understands Slovene better than other open-source models of similar size (except for GaMS), and has a larger context window than previous generations of GaMS models. We perform the post-processing in two steps. In the first step, Gemma 3 is asked to classify each page as either content or boilerplate. This step is necessary because there are many empty or boilerplate pages at the beginning of each book. Additionally, documents could contain pages that include only images or advertisements, or the entire page was converted into a single artefact during an OCR. Such pages need to be removed before merging.

After the classification, the page pairs are merged into a single document. We achieve this by iterating through the pages and sending the last paragraph of the current page and the first paragraph of the next page into Gemma 3. Gemma 3 is then asked to merge the pages based on one of the following scenarios:
\begin{itemize}
    \item The last paragraph of the current page represents the page footer. In this case, this paragraph is removed in the merged version.
    \item The first paragraph of the next page represents the page header. In this case, that paragraph is removed in the merged version.
    \item The last paragraph of the current page ends with a hyphenated word that continues in the next page. In this case, the paragraphs are merged by concatenating the words without the "-" sign.
    \item The last paragraph of the current page and the first paragraph of the next page represent the same paragraph. In this case, they are merged into a single paragraph.
    \item The last paragraph of the current page and the first paragraph of the next page represent different paragraphs. In this case, the pages are merged using a line-break.
\end{itemize}

\subsubsection{Filtering stage}

In the final stage, markdown documents are put through a series of NeMo Curator~\footnote{\url{https://github.com/NVIDIA-NeMo/Curator}} modifiers. All documents are sent through the following filters:
\begin{itemize}
    \item \textbf{Image remover} removes markdown embedded images.
    \item \textbf{Newline normalizer} truncates consecutive newline characters to at most two.
    \item \textbf{Unicode reformatter} corrects improperly encoded unicode characters.
    \item \textbf{ČŠŽ corrector}: replaces the wrong forms of Slovene diacritic characters č, š, and ž (such as ˇc) with the correct ones.
\end{itemize}

In the case of Nanonets, we add two additional filters to remove the artefacts produced by the model. The first filter removes paragraphs that are too long (paragraphs longer than 15,000 characters), as this indicates either an artefact or repeated words or sentences (the model started cycling). The second filter removes repeated paragraphs. If the paragraph repeats more than 100 times in a document, this suggests cycling, and only the first occurrence is kept.

\subsection{Final data preparation}
\label{sec:cpt_data_prep}

We performed additional data processing before the training to improve both the quality and efficiency of the training. Using this data step, we set the length of each training example to exactly the sequence length used in each CPT stage. This is necessary when using the NeMo framework~\footnote{https://github.com/NVIDIA-NeMo/NeMo}, as otherwise the framework itself can split training documents across multiple training examples, i.e., if a document ends before the sequence length, the framework takes the start of the next document, splitting that document randomly. Besides inadequate context, this causes problems with input sequences, as the next training instance does not start with the Beginning Of Sequence (BOS) token, expected by most LLMs. For Gemma-type models, this is especially problematic,  as their performance degrades substantially when the BOS token is not included. To avoid the situation, we used the following procedure.

First, we tokenized each document and split documents that are longer than the target sequence length into smaller units. Units differ between corpora and are either sentences, paragraphs, or sections. We then merge consecutive units that do not exceed the context window length into subdocuments. This is followed by a heuristics-based sequence packing, where we merge split documents and subdocuments into a single training example that does not exceed the context window length. Finally, all training examples, shorter than the context window, are padded with the End Of Sequence (EOS) tokens.

\section{Supervised fine-tuning}
\label{sec:sft}
After continual pretaining, the next phase of LLM adaptation to a specific language is the supervised fine-tuning (SFT) phase that improves the final performance of the model on various tasks. While the CPT phase builds the model's knowledge and can significantly enhance the language and understanding of the culture, the SFT phase enables the model to express its knowledge. The impact of the SFT phase was evident in the previous generation of GaMS models, where the difference between GaMS-9B-Instruct~\citep{gams_9b} and GaMS-9B-Instruct-Nemotron~\citep{gams_9b_nemotron} is only in the SFT data, but the difference in their ranking on the Slovene LLM Arena is significant. While GaMS-9B-Instruct ranks among the worst-performing models, GaMS-9B-Instruct-Nemotron is one of the top models, beating commercial GPT-4o and Gemini-2.0-Flash.

For GaMS3, we split the instruction tuning into two SFT stages: general instruction tuning and chat tuning. We based the general instruction tuning stage on the GaMS-Instruct dataset and the chat tuning stage on the GaMS-Nemotron-Chat dataset. Both datasets were specifically designed for the instruction tuning of Slovene models. We describe the generation of the GaMS-Instruct dataset in Section \ref{sec:gams_instruct} and the preparation of the GaMS-Nemotron-Chat dataset in Section \ref{sec:gams_nemotron_chat}.

The idea of the general instruction tuning stage is to train the model on many different task types and topics. After this stage the model should be useful for downstream tasks. We used the following combination of datasets at this stage:
\begin{itemize}
    \item \textbf{GaMS-Instruct}: we combined the newly created general-purpose GaMS-Instruct dataset consisting of ClosedQA, OpenQA, and Writing tasks with the digital humanities based GaMS-Instruct-DH 1.0~\citep{gams_inst_dh} dataset.
    \item \textbf{GaMS-Lex} is a collection of questions and answers about Slovene lexicography, derived from multiple sources, such as questions based on ssj500k~\citep{ssj500k}, questions from Jezikovna svetovalnica~\footnote{\url{https://svetovalnica.zrc-sazu.si}}, and manually written questions during prompathon.
    \item \textbf{Nemotron-Post-Training-v2}~\citep{nemotron_post_training}: We used the STEM and Math splits of this dataset, as they include topics not well covered by other datasets in our mix. We translated the subset of examples to Slovene using Gemini-2.5-Flash. We included both English and Slovene examples; however, we ensured that the English examples were different from the ones translated to Slovene.
    \item \textbf{Slovenian Code Feedback}~\citep{sl_code} is a translated code dataset based on English CodeFeedback and similar datasets. The dataset was translated to Slovene using Gemini-2.5-Flash and contains different task types and programming languages. Code datasets are generally considered beneficial for LLM training as they contain structured data with many implementations of various algorithms.
\end{itemize}

The number of examples in each dataset is shown in Table \ref{tab:sft_instruct_stats}. We used the $ 90 / 10 $ train validation split ratio for most of the datasets. The exceptions are the Slovene part of the Nemotron datasets, where we downsampled the data to prevent overfitting to its specific task types, and GaMS-Lex, which we used only for training due to its limited data size.

\begin{table}[hbt!]
\begin{threeparttable}
\caption{Number of training and validation examples per dataset in the general instruction tuning stage.}
\label{tab:sft_instruct_stats}
\begin{tabular}{l l r r}
\hline
\headrow Dataset & Language & Number of training examples & Number of validation examples \\
\hline
GaMS-Lex & Slovene & 1,884 & 0 \\
GaMS-Instruct-ClosedQA & Slovene & 10,825 & 1,202 \\
GaMS-Instruct-OpenQA & Slovene & 28,704 & 3,189 \\
GaMS-Instruct-Writing & Slovene & 9,056 & 1,006 \\
GaMS-Instruct-DH 1.0 & Slovene & 9,135 & 1,015 \\
Nemotron-SFT-v2-Math-En & English & 9,000 & 1,000 \\
Nemotron-SFT-v2-Math-Sl & Slovene & 10,000 & 1,000 \\
Nemotron-SFT-v2-STEM-En & English & 9,000 & 1,000 \\
Nemotron-SFT-v2-STEM-Sl & Slovene & 10,000 & 1,000 \\
SlCode & Slovene & 10,000 & 1,000 \\
\hline
\textbf{Total} &  & \textbf{107,604} & \textbf{11,412} \\
\hline
\end{tabular}
\end{threeparttable}
\end{table}

In the chat tuning stage, the model is trained to perform better as a general purpose chat model. The GaMS-Nemotron-Chat dataset showed promising results in this area with the previous generation of GaMS models. The dataset guides the model to be more user-friendly, making it more helpful, elaborate, and able to better follow the requested response format. To make the model safer, we add a small set of safety prompts to this stage. The safety prompts were written by humans, and we plan to gradually increase this dataset in the future. The dataset sizes for the chat tuning stage are shown in Table \ref{tab:sft_chat_stats}. We used a 90/10 training-validation split for the GaMS-Nemotron-Chat dataset, while all safety examples were included in the training set due to the small dataset size.

\begin{table}[hbt!]
\begin{threeparttable}
\caption{Number of training and validation examples per dataset in the chat tuning stage.}
\label{tab:sft_chat_stats}
\begin{tabular}{l l r r}
\hline
\headrow Dataset & Language & Number of training examples & Number of validation examples \\
\hline
GaMS-Safety & Slovene & 459 & 0 \\
GaMS-Nemotron-Chat & Slovene, English & 88,126 & 9,791 \\
\hline
\textbf{Total} &  & \textbf{88,585} & \textbf{9,791} \\
\hline
\end{tabular}
\end{threeparttable}
\end{table}

\subsection{GaMS-Instruct dataset}
\label{sec:gams_instruct}

As less-resourced languages such as Slovene lack large-scale, high-quality datasets for instruction tuning of LLMs, we created GaMS-Instruct, a comprehensive instruction tuning dataset for Slovene, following state-of-the-art construction guidelines that emphasize the importance of balanced datasets covering diverse tasks and topics. The dataset comprises 53,983 instruction-response pairs and is designed to support the tuning of Slovene LLMs. 

\subsubsection{Dataset construction}

We organized the dataset into three main parts, each corresponding to a common type of prompts submitted by users. The overall distribution and size of instruction-response pair types are based on recent findings and are shown in Table \ref{tab:sft_instruct_stats}. The dataset contains the following task types: 

\begin{itemize}
  \item \textbf{Open questions}, covering a broad range of topics and subtopics;
  \item \textbf{Closed questions}, generated on the basis of Slovene source texts;
  \item \textbf{Writing tasks}, targeting predefined text types.
\end{itemize}

\paragraph{Open question answering}
Preliminary experiments revealed that unconstrained open-question generation resulted in an uneven topic distribution. To address this, we manually selected topics for generating open-question tasks. To ensure broad and balanced coverage across domains, topic selection was based on the IPTC subject taxonomy for media~\footnote{\url{https://iptc.org/standards/media-topics}}, Wikipedia’s category system \footnote{\url{https://en.wikipedia.org/wiki/Wikipedia:Contents/Categories}}, and Encyclopaedia Britannica’s online categorization~\footnote{\url{https://www.britannica.com/}}. In total, we produced 20 broad categories that span a wide range of domains. Each category was further subdivided into 30 more specific subtopics. Each subtopic was reused multiple times during instruction generation to ensure a sufficient number of examples per topic. 

\paragraph{Closed question answering}

Closed-question tasks were generated from three types of existing Slovene textual sources, with  answers extracted directly from the source text:

\begin{itemize}
  \item \textbf{Trendi}~\citep{trendi}, a corpus of Slovene news and web content, with 3,000 examples;
  \item \textbf{KAS}~\citep{kas}, a corpus of academic theses in Slovene, with 3,000 examples;
  \item \textbf{Slovene Wikipedia}~\citep{wikipedia_markdown}, with 6,028 examples.
\end{itemize}

\paragraph{Writing tasks}
We designed the distribution of writing tasks so that it would reflect the frequency with which different text types are requested by LLM users. We divided possible writing tasks into three categories based on their frequency:

\begin{itemize}
  \item \textbf{High-frequency taks} (e.g., emails, essays, and stories), generated 400 times each;
  \item \textbf{Medium-frequency tasks} (e.g., manuals, recipes, and speeches), generated 140 times each;
  \item \textbf{Writing tasks} (e.g., greeting cards, menus, or petitions), generated 25 times each.
\end{itemize}

In addition to generating texts of a given type, we also produced tasks covering activities similar to writing such as rewriting, editing, simplifying, etc.

\subsubsection{Instruction generation}

To generate prompts for each of the described domains, we adapted the Self-Instruct framework for multilingual instruction generation~\citep{wang2023self}, with a specific focus on the Slovene language. Unlike the original approach, which relies on a fixed set of manually written English seed tasks, our method generates instructions directly from topic-specific inputs or source documents, without predefined seed examples.
To improve the instructional diversity of the dataset, approximately half of the dataset was generated using Gemini 2.0 Flash, and the other half was generated using GPT-4o. A structured prompting system dynamically created both system and user prompts, with selected prompt templates adopted from the NeMo Curator toolkit. 
To ensure novelty and reduce redundancy, we filtered newly generated instructions using ROUGE-L similarity against previously generated instructions. For tasks that did not already include an answer, such as writing tasks, the model was additionally prompted to produce a valid response, which resulted in complete instruction-response pairs suitable for instruction tuning.

\subsection{GaMS-Nemotron-Chat}
\label{sec:gams_nemotron_chat}

To improve the conversational quality of our model, we develop the GaMS-Nemotron-Chat dataset~\citep{llmarena2025petric}.
We created it by translating English samples from the Nemotron-Post-Training-Dataset-v1~\citep{nathawani2025nemotron}.
This dataset contains responses to real-world user prompts from LMSYS Chat 1M~\citep{ds-lmsyschat1m-zheng2023lmsyschat1m}. The responses in the dataset are regenerated by the Qwen3-235B-A22B~\citep{qwen3-235b-a22b-yang2025qwen3} model, which achieves good performance on benchmarks such as ArenaHard-v2~\citep{arena-hard-benchbuilder-pipeline-li2025from}.
To ensure diversity and reduce redundancy, we started with the LMSYS Chat 1M Clean subset~\citep{ds-lmsys-clean-openleecher2024} containing approximately 273,000 examples.
We then applied MinHash LSH filtering~\citep{minhash-lsh-indyk1998approximate} within each category (e.g., coding, creative writing) using a similarity threshold of 0.65 and 256 permutations, resulting in a deduplicated selection of approximately 80,000 distinct, high-quality examples.

We translated the deduplicated examples into Slovene using the GaMS-27B-Instruct model, employing a custom prompt template, shown in \ref{app:prompt-gams-27b-translation}, to preserve the conversation structure (i.e., \texttt{<user>} and \texttt{<assistant>} tags).
To ensure the quality of the synthetic data, we implemented a length-ratio filter.
We retained only the translations where the ratio of the translated text length to the source text length fell within the range of $[0.73, 1.35]$.
This heuristic proves effective in preventing hallucinations and incomplete translations, which were frequently observed outside this interval.
During the initial phases, we identified two critical issues: the model frequently refers to itself as "Qwen" due to the source data, and it responds in Slovenian when prompted in English.
To correct the identity issue, we identified 694 conversations containing references to "Qwen" and regenerated the responses using a prompt, shown in \ref{app:prompt-gams-identity}. This approach explicitly established the model's identity as GaMS.

To mitigate the language mixing issue and preserve the model's ability to respond in English, we augmented the 80,000 Slovene translations with approximately 20,000 original English examples from the Nemotron-Post-Training-Dataset-v1, ensuring no overlap with the translated subset.
The final GaMS-Nemotron-Chat dataset thus consists of roughly 100,000 examples, with an $ 80/20 $ ratio between Slovene and English.

\section{Training setup}
\label{sec:training}
In this section, we describe the technical details of adapting the LLM to the less-resourced Slovene language. 
We took Gemma 3 12B (pre-trained version) as our starting point. We performed CPT using NeMo 2.0 framework (we used version 25.07 of the official NVIDIA NeMo container~\footnote{\url{https://catalog.ngc.nvidia.com/orgs/nvidia/containers/nemo?version=25.07}}). We chose NeMo as it supports Gemma 3 and offers various optimizations and parallelisms. In combination with an official container, it enables efficient scaling across multiple GPUs and nodes. For the SFT, we switched to Transformers (version \texttt{4.57.3}) and DeepSpeed~\footnote{\url{https://huggingface.co/docs/transformers/deepspeed}} (version \texttt{0.18.2}) as we observed some issues with the quality of Gemma 3 SFT in NeMo. As this stage is less computationally intensive, it can be performed in a less-optimized environment.

During the CPT stage, we used the following optimization techniques and parallelisms:
\begin{itemize}
    \item \textbf{Tensor parallelism (TP)} splits each linear layer and attention heads across multiple GPUs.
    \item \textbf{Sequence parallelism} splits layers such as dropout and layernorm across TP ranks.
    \item \textbf{Activation recomputation} (also known as activation checkpointing): instead of storing all the activations for the backward pass, it stores only the activations before each attention layer and recomputes activations inside the layer during the backward pass. This reduces vRAM usage.
\end{itemize}

For the SFT, we used DeepSpeed ZeRO stage 2. The hyperparameters for all stages are shown in Table \ref{tab:training_hyperparameters}. We used \texttt{bf16} precision in all training stages.

\begin{table}[hbt!]
\begin{threeparttable}
\caption{Training hyperparameters across the CPT and SFT stages. During the base CPT stage, we used a ramp-up batch size, where the batch size increased from 128 to 192 after 961 steps. It then increased again from 192 to 256 after another 600 steps. Both SFT stages were performed for 3 epochs; however, the checkpoints after the second epoch were selected based on the validation loss. Hyperparameters for both SFT stages were found using grid search.}
\label{tab:training_hyperparameters}
\resizebox{\textwidth}{!}{
\begin{tabular}{l|ccc|cc}
\toprule
\headrow & \multicolumn{3}{c|}{CPT} & \multicolumn{2}{c}{SFT} \\
\headrow Hyperparameter & Parallel Alignment & Base CPT & Long CPT & Instruction tuning & Chat tuning \\
\midrule
Model Parallelism & TP 8 & TP 8 & TP 8 & DeepSpeed ZeRO Stage 2 & DeepSpeed ZeRO Stage 2 \\
Data Parallelism  & 64 & 64 & 16 & 8 & 8 \\
Batch Size &
128 &
128 $\rightarrow$ 192 $\rightarrow$ 256 &
64 &
64 &
64 \\
Micro Batch Size & 1 & 1 & 1 & 8 & 8 \\
LR Scheduler & Cosine with warmup & Cosine with warmup & Constant with warmup & Cosine with warmup & Cosine with warmup \\
Min LR & $5\times10^{-7}$ & $5\times10^{-7}$ & / & $1\times10^{-6}$ & $1\times10^{-6}$ \\
Max LR & $5\times10^{-6}$ & $5\times10^{-6}$ & $5\times10^{-6}$ & $5\times10^{-6}$ & $5\times10^{-6}$ \\
Warmup Steps & 150 & 1000 & 500 & 1000 & 1000 \\
Constant Steps & 200 & 1000 & / & 0 & 0 \\
Epochs &
1 &
1 &
1 &
2 &
2 \\
\bottomrule
\end{tabular}}
\end{threeparttable}
\end{table}

We used the following hardware for training:
\begin{itemize}
    \item \textbf{LEONARDO Booster} is a EuroHPC supercomputer, where its Booster partition has 3200 4xA100 64GB nodes. The GPUs on a single node are connected through 600 GB/s NVLink, and the nodes are connected through 2x200 Gb/s Infiniband. We managed to scale the training across 128 nodes (the hard limit set by the cluster is 256 nodes). We used the LEONARDO Booster for the parallel alignment stage and a large part of the base CPT stage. We spent approximately 140k GPU hours on LEONARDO for training.
    \item \textbf{Internal cluster FRIDA}: we used a single DGX 8xB200 180GB node and a single DGX 8xH100 80GB node belonging to the internal FRIDA cluster. We used these two nodes for a small portion of the base CPT stage, when we ran out of resources on LEONARDO. We spent approximately 120 B200 GPU hours and 960 H100 GPU hours for training.
    \item \textbf{NVIDIA DGX Cloud Lepton} is a unified AI platform that connects developers to tens of thousands of GPUs from a global network of cloud providers. We were given access to 16 8xH200 141GB nodes. We used Lepton for finalizing the base CPT stage and all remaining stages. We were allocated 40k GPU hours.
\end{itemize}

\section{Evaluation}
\label{sec:evaluation}

We evaluated GaMS3 with three scenarios. In Section \ref{sec:slo_llm_eval}, we present the evaluation on the Slovenian-LLM-Eval suite of benchmarks, which tests the general knowledge of the model.  We present an evaluation of a specific generative task in Section \ref{sec:translation_eval}, where we evaluate the model's performance on English-to-Slovene translation. Finally, we evaluate the model's generative capabilities using the Slovene-LLM-Arena in Section \ref{sec:llm_arena}.

\subsection{Slovenian-LLM-Eval}
\label{sec:slo_llm_eval}

Slovenian-LLM-Eval~\citep{slo_eval} is a collection of benchmarks that are part of the LM Evaluation Harness~\citep{lm_eval}. Slovene versions of English benchmarks were machine translated into Slovene. A part of them was translated by Aleksa Gordić and his community using Google's translation service and then refined using GPT-4o. The other part was translated using DeepL as a part of the European LLM Leaderboard~\citep{euro_lm_eval}. We used the following set of benchmarks: ARC-Easy, ARC-Challenge, BoolQ, GSM8K, HellaSwag, OpenBookQA, PIQA, TruthfulQA and Winogrande. Additionally, we asked native Slovene speakers to manually inspect and correct ARC-Easy, ARC-Challenge, OpenBookQA and Winogrande.

We compared GaMS3 12B with multilingual open-source models of a similar size. We used instruction-tuned versions of all models. Per-benchmark results are shown in Table \ref{tab:slo_eval_benchmarks}. GaMS3 performs well across benchmarks compared to Zlatorog and GaMS-9B. It performs slightly worse than GaMS-27B, which is twice as large. Compared to Gemma 3 12B, GaMS3 performs significantly better on almost all benchmarks. The only exceptions are GSM8K and TruthfulQA. We hypothesize that the reason for the worse performance on those is the few-shot evaluation. This points to the need to improve the performance of GaMS3 on longer inputs. We could also expand our SFT datasets to include few-shot tasks as well. Another interesting finding is that Slovene-specialized models outperform practically all open-source models on benchmarks that were manually corrected by native Slovene speakers. This only emphasizes the need for manual correction of other benchmarks as well.

\begin{table}[hbt!]
\begin{threeparttable}
\caption{Per-benchmark Slovenian-LLM-Eval results for Slovene specialized (top) and multi-lingual open-source models (bottom). Higher scores are better. The score of the best performing model for each benchmark is marked with \textbf{bold} and the score of the second best model is \underline{underlined}. Acc. stands for accuracy and EM stands for strict exact match. The column n-shot shows the number of solved examples that were added to the prompt. Benchmarks that were manually corrected by native Slovene speakers are marked with †.}
\label{tab:slo_eval_benchmarks}
\resizebox{\textwidth}{!}{
\begin{tabular}{l|lr|ccccc}
\toprule
\headrow Benchmark & Metric & n-shot &
GaMS3-12B &
GaMS-9B-Nemotron &
GaMS-27B-Nemotron &
Zlatorog-12B \\
\midrule
ARC Challenge † & Acc. & 0 & \underline{0.5265} & 0.5171 & \textbf{0.5444} & 0.4676 \\
ARC Easy †      & Acc. & 0 & \underline{0.7744} & 0.7546 & \textbf{0.7875} & 0.7079 \\
BoolQ           & Acc. & 0 & 0.8523 & 0.8471 & \underline{0.8618} & 0.8321 \\
GSM8K           & EM & 5 & 0.6892 & 0.6277 & 0.6892 & 0.5921 \\
HellaSwag       & Acc. & 0 & 0.5111 & 0.5140 & \textbf{0.5428} & \underline{0.5373} \\
OpenBookQA †    & Acc. & 0 & 0.3940 & \textbf{0.4080} & 0.3980 & \underline{0.4060} \\
PIQA            & Acc. & 0 & 0.7149 & 0.7062 & \textbf{0.7252} & \underline{0.7247} \\
TruthfulQA MC1  & Acc. & 6 & 0.3807 & 0.3415 & 0.3672 & 0.3647 \\
TruthfulQA MC2  & Acc. & 6 & 0.5396 & 0.5307 & 0.5314 & 0.5383 \\
Winogrande †    & Acc. & 0 & 0.7056 & \underline{0.7190} & \textbf{0.7206} & 0.7017 \\
\bottomrule
\end{tabular}}
\resizebox{\textwidth}{!}{
\begin{tabular}{l|lr|cccccccc}
\toprule
\headrow Benchmark & Metric & n-shot &
Gemma 3 12B &
Gemma 3 27B &
EuroLLM-22B &
Qwen3-30B-A3B &
Apertus-8B &
Bielik-11B-v3.0 \\
\midrule
ARC Challenge † & Acc. & 0 & 0.4514 & 0.5137 & 0.5026 & 0.4386 & 0.4556 & 0.4923 \\
ARC Easy †      & Acc. & 0 & 0.6936 & 0.7475 & 0.7433 & 0.6427 & 0.7071 & 0.7231 \\
BoolQ           & Acc. & 0 & 0.8526 & \textbf{0.8630} & 0.8278 & 0.8590 & 0.8330 & 0.8596 \\
GSM8K           & EM & 5 & \underline{0.7430} & \textbf{0.8006} & 0.5148 & 0.7202 & 0.4375 & 0.7339 \\
HellaSwag       & Acc. & 0 & 0.4728 & 0.5235 & 0.4931 & 0.4256 & 0.4888 & 0.5212 \\
OpenBookQA †    & Acc. & 0 & 0.3520 & 0.3700 & 0.3880 & 0.2860 & 0.3420 & 0.3780 \\
PIQA            & Acc. & 0 & 0.6616 & 0.7106 & 0.6904 & 0.6246 & 0.6893 & 0.6937 \\
TruthfulQA MC1  & Acc. & 6 & \underline{0.3978} & \textbf{0.4076} & 0.3537 & 0.3953 & 0.3917 & 0.3794 \\
TruthfulQA MC2  & Acc. & 6 & 0.5827 & \textbf{0.5881} & 0.5229 & \underline{0.5845} & 0.5574 & 0.5517 \\
Winogrande †    & Acc. & 0 & 0.6559 & 0.6717 & 0.6551 & 0.6204 & 0.6496 & 0.6922 \\
\bottomrule
\end{tabular}}
\end{threeparttable}
\end{table}

For easier comparison between models across all datasets, we rank the models from 1 (best) to 10 (worst) for each benchmark metric and compute the average across all benchmarks. The results are shown in Table \ref{tab:slo_eval_leaderboard}. GaMS3 ranks only behind the two times larger GaMS-27B-Nemotron and Gemma 3 27B. It beats Gemma 3 12B significantly, showing a clear benefit of Slovene adaptation. GaMS3 also outperforms GaMS-9B-Nemotron and Zlatorog, which are the Slovene specialized models of similar size. Among multi-lingual models, the best performing models are Gemma 3 27B and Bielik-11B, which is surprising, as the latter is predominantly trained on Polish. On the other hand, EuroLLM and Apertus, which should support all EU languages, perform quite badly.

\begin{table}[hbt!]
\begin{threeparttable}
\caption{Slovene-LLM-Eval leaderboard based on average model rank across benchmarks. The lower rank is better.}
\label{tab:slo_eval_leaderboard}
\begin{tabular}{l c}
\toprule
\headrow Model & Average rank \\
\midrule
GaMS-27B-Nemotron & 3.05 \\
Gemma 3 27B & 3.20 \\
\textbf{GaMS3-12B} & 4.25 \\
Bielik-11B & 5.00 \\
GaMS-9B-Nemotron & 5.20 \\
Zlatorog-12B & 5.60 \\
Gemma 3 12B & 6.30 \\
Qwen3-30B-A3B & 7.30 \\
EuroLLM-22B & 7.50 \\
Apertus-8B & 7.60 \\
\bottomrule
\end{tabular}
\end{threeparttable}
\end{table}

\subsection{English to Slovene translation}
\label{sec:translation_eval}

To evaluate GaMS3-12B-Instruct's English-to-Slovene translation abilities, we used an upgraded version of our custom benchmark \citep{gams_translator}. We upgraded the benchmark with an additional dataset and metrics of interest. The original benchmark includes articles from CC-News and English Wikipedia. For this evaluation, we added examples from the GaMS-Nemotron-Chat dataset, which are used to evaluate markdown formatting preservation. Such evaluation is necessary for models used for chat data translation, which is markdown-formatted, as a potential formatting error could lead to misleading chat tuning of the target models.

Similarly to \citet{gams_translator}, we translated English documents into Slovene using the simple \emph{Translate the following English document to Slovene} prompt. We computed the following metrics: COMET for semantic similarity, language error to detect responses in the wrong language, and truncation error to detect responses that are too short. The COMET score is calculated for each dataset individually with the reference-less Direct Assessment (DA) model from CometKiwi~\citep{cometkiwi}. The language detection is performed with a pre-trained language identification model from the FastText library~\citep{fasttext_lid}. The truncation error is detected using a heuristic. If the ratio between the number of characters in the translated and original text is below $ 0.7 $, the example is flagged as truncated. Finally, we measured the percentage of examples with markdown formatting errors using Gemma-3-27b-it model~\citep{gemma}, which is given the source and the translated text and is instructed to classify whether the formatting matches perfectly (see \ref{app:markdown-check} for the full prompt).

We compared GaMS3 with the previous generation of GaMS models, the specialized GaMS translator, commercial Gemini-2.5-flash, and the open-source EuroLLM. The results are shown in Table \ref{tab:translation_results}. GaMS3-12B-Instruct matches the overall COMET performance of GaMS-27B-Instruct despite having fewer parameters. It is worth mentioning that GaMS-27B-Instruct was also trained on a subset of examples from the RSDO translation dataset during the SFT, and GaMS3 was not. GaMS3 also exhibits substantially lower error rates for language and formatting errors. We attribute this gap primarily to the limitations in the training data. GaMS-27B-Instruct was continually pre-trained on a substantially lower amount of tokens (between 15 and 20 B), and it appears undertrained for its scale, while GaMS3 benefits from CPT on a larger amount of Slovene text. Additionally, GaMS-Nemotron-Chat was not part of the base GaMS-27B-Instruct training, which was instruction-tuned on a significantly lower amount of data (around 25k examples), leading to more formatting errors. We also need to mention that GaMS3 was pretrained on Wikipedia translated with GaMS-9B-SFT-DPO-Translator, resulting in higher COMET scores on the Wikipedia dataset.

Compared to Gemma 3, GaMS3 achieves significantly higher COMET scores across all three datasets, showing that Slovene specialization boosted the semantic similarity of translations. As expected, it  makes fewer language errors. The benefit of Slovene specialization is also reflected in the amount of markdown errors. However, GaMS3 makes more truncation errors than both Gemma 3 models. This suggests that we shall focus on improving the long-context SFT data in the future. While GaMS models exhibit slightly lower performance than Gemini, it should be noted that Gemini is significantly larger. Furthermore, Gemini is one of the leading commercial models for translation, achieving results comparable to those of specialized systems, such as DeepL. Compared to EuroLLM, GaMS3 performs better in both terms of COMET score and error rates, even though EuroLLM was trained on a large amount of parallel data and also saw a significant number of translation examples during the SFT stage.

\begin{table}[hbt!]
\begin{threeparttable}
\caption{English to Slovene translation results. The overall COMET score is computed as the average over the three datasets. The metrics, for which the higher score is better, are marked with ($\uparrow$), and the metrics for which the lower score is better are marked with ($\downarrow$). The best score for each metric is in \textbf{bold}, and the second best result is \underline{underlined}.}
\label{tab:translation_results}
\centering
\resizebox{\textwidth}{!}{
\begin{tabular}{l|cccc|ccc}
\toprule
\headrow & \multicolumn{4}{c|}{COMET ($\uparrow$)} & & & \\
\headrow Model & Overall & CC-News & Nemotron-Chat & Wikipedia & Lang. err. ($\downarrow$) & Trunc. err. ($\downarrow$) & Markdown err. ($\downarrow$) \\
\midrule
Gemini-2.5-flash & \textbf{0.717982} & \textbf{0.702981} & \textbf{0.697498} & \textbf{0.753924} & \textbf{0.35\%} & 0.42\% & \textbf{3.70\%} \\
GaMS-9B-SFT-DPO-Translator & \underline{0.708042} & \underline{0.702903} & 0.679462 & \underline{0.742583} & \underline{0.91\%} & 0.28\% & \underline{18.28\%} \\
GaMS-27B-Instruct & 0.701284 & 0.686480 & \underline{0.680014} & 0.730733 & 27.28\% & 5.36\% & 62.07\% \\
GaMS3-12B-Instruct & 0.700226 & 0.684848 & 0.674523 & 0.741844 & 1.11\% & 1.83\% & 32.61\% \\
GaMS-9B-Instruct & 0.693659 & 0.685006 & 0.673394 & 0.723470 & 13.50\% & 4.83\% & 33.15\% \\
EuroLLM-9B-Instruct & 0.689321 & 0.668084 & 0.670723 & 0.729227 & 8.97\% & 1.89\% & 35.08\% \\
Gemma-3-27b-it & 0.668378 & 0.643716 & 0.660861 & 0.704403 & 5.93\% & \textbf{0.0\%} & 39.17\% \\
Gemma-3-12b-it & 0.613083 & 0.602837 & 0.619401 & 0.621018 & 16.48\% & \underline{0.19\%} & 57.32\% \\
\bottomrule
\end{tabular}}
\end{threeparttable}
\end{table}

\subsection{Slovene-LLM-Arena}
\label{sec:llm_arena}

Inspired by popular Chatbot arenas such as LMSYS~\citep{lmsys_arena,ds-lmsyschat1m-zheng2023lmsyschat1m}, we introduced the Slovene LLM Arena with the goal of evaluating models on Slovene prompts by Slovene users. The user is first asked to write a prompt and gets responses from two different models. The model names are hidden from the user, and the models' responses are labelled as \texttt{Model A} and \texttt{Model B}. The user can either rank the responses or continue the conversation. When ranking the responses, the user can choose between 4 options: model A is better, model B is better, tie, or both responses are bad. The model names are revealed to the user only after the ranking is complete. The leaderboard is computed using the ELO score~\footnote{\url{https://en.wikipedia.org/wiki/Elo_rating_system}} known from chess.

We compared several multi-lingual and commercial models in the arena. A leaderboard snapshot from January 13th, 2026 (the leaderboard is constantly updating) is shown in Table \ref{tab:slo_arena}. When analyzing the results, it has to be taken into account that some models have a low number of votes and the ELO Scores could be misleading. However, combining them with the win rate is informative. We can see that GaMS3 performs slightly worse than the GaMS-Nemotron models. The main reason for the worse performance is that the Slovene generated by GaMS3 appears to be slightly more machine-translated than the Slovene from the other two GaMS models. This shows the need for curation and improvement of our machine-translated dataset, especially the GaMS-Nemotron-Chat dataset and the datasets in the long CPT phase.

\begin{table}[hbt!]
\begin{threeparttable}
\caption{The leaderboard of top 15 models based on ELO Score on Slovene-LLM-Arena on the 13th of January 2026. Win rate is computed as a ratio between the number of wins and sum of wins and looses (ties are discarded).}
\label{tab:slo_arena}
\centering
\begin{tabular}{r l c c c}
\toprule
\headrow Rank & Model & ELO Score & Win Rate & Total Votes \\
\midrule
1  & Gemini-2.5-Pro              & 1100 & 78.3\%  & 181 \\
2  & GaMS-27B-Instruct-Nemotron  & 1062 & 66.3\%  & 240 \\
3  & Gemma-3-27b-it              & 1049 & 66.1\%  & 177 \\
4  & GaMS-9B-Instruct-Nemotron   & 1043 & 64.2\%  & 129 \\
5  & Gemini-2.0-Flash            & 1036 & 61.0\%  & 156 \\
6  & GPT-4o                      & 1030 & 59.0\%  & 181 \\
7  & GaMS3-12B-Instruct          & 1025 & 61.2\%  & 95  \\
8  & Kimi-K2-Instruct            & 1015 & 58.5\%  & 52  \\
9  & Gemini-3-Pro-Preview        & 1008 & 83.3\%  & 6   \\
10 & DeepSeek-R1                 & 1005 & 51.2\%  & 166 \\
11 & GPT-5                       & 1005 & 75.0\%  & 5   \\
12 & Gemini-1.5-Pro              & 1004 & 52.1\%  & 159 \\
13 & DeepSeek-R1-0528            & 1003 & 51.2\%  & 63  \\
14 & GPT-5-Mini                  & 1002 & 100.0\% & 1   \\
15 & Mistral-Large-3-675B        & 1000 & 50.0\%  & 4   \\
\bottomrule
\end{tabular}
\end{threeparttable}
\end{table}

Compared to commercial models, GaMS3 performs similarly to GPT-4o and Gemini-2.0-Flash, which is a very good result, given the size and resource advantage of these two models. All GaMS models are surpassed by Gemini-2.5-Pro, and we believe they will be surpassed by Gemini-3-Pro and GPT-5 as well once they reach a sufficient number of votes. However, this is expected as these are the latest and best-performing models from the leading model builders. We observe that commercial models outperform the GaMS3 model on more challenging tasks, i.e., tasks that require reasoning, which suggests that we should incorporate this capability into our future models.

Compared to open-source models, GaMS3 is beaten only by Gemma 3 27B, a very strong multi-lingual model, and the GaMS-Nemotron models. Gemma 3 12B, which was the base model for GaMS3, did not make it into the top 15 and currently ranks only \textbf{25}th with an ELO Score of \textbf{979} and the win rate of \textbf{42.6\%} after \textbf{168} votes. Zlatorog, another specialized model for Slovene, is ranked \textbf{23rd} with an ELO Score of \textbf{986} and a win rate of \textbf{45.2\%} after \textbf{166} votes. These results show a clear benefit of our Slovene specialization procedure.

\section{Conclusion}
\label{sec:conclusion}

We presented the development of GaMS3-12B, an open-source LLM specialized for Slovene. We provided a detailed description of the procedure and data preparation, which allowed us to continually pretrain and supervisedly fine-tune GaMS3, starting from the Gemma 3 12B model. We provided a detailed evaluation of the GaMS3 model in three different scenarios: a suite of Slovene benchmarks, Slovenian-LLM-Eval (general knowledge), a downstream generative task (machine translation), and general chat performance (Slovene LLM Arena).

The main drawback of our approach appears to be its reliance on machine-translated data, particularly during the SFT stage. As a result, the model's Slovene is not as natural as it could be. We plan to address this problem in the future by reducing the amount of machine-translation data and improving its quality. Areas where our model could improve are the difficult tasks, both in terms of long contexts and logical reasoning. We plan to address this issue by increasing the amount of long context data and adding reasoning data to our SFT dataset. Additionally, we focused only on a language modality, despite the multi-modality of Gemma 3. We plan to address this issue in the future by joining forces with the SVILA (Slovene VLM based on Gemma 3) model. GaMS3 will also be trained for tool-calling in the future, enabling the agentic approaches.

Besides the open-source model, the most significant contribution is a detailed description of our training approach and data preparation. We demonstrated that this approach can significantly enhance the base model's performance in less-resourced languages. Our GaMS3 model outperforms Gemma 3 12B in all evaluation scenarios and matches the performance of much larger commercial models such as GPT-4o and Gemini-2.0-Flash in the Slovene LLM Arena. Collaborating with similar initiatives, we believe that our work could significantly contribute to the development of similar open-source models for other less-resourced languages.

\paragraph{Acknowledgments}
We thank everyone who contributed to data collection and preparation, which enabled us to train our model. A special thanks goes to Jaka Čibej and his team for manual correction of benchmarks, Tomaž Savodnik for converting the Wikipedia to markdown format and crawling the universities' repositories, Miha Malenšek for preparation of Slovene legal and medical data, Alina Machidon for preparation of the Slovene code dataset, the National and University Library for sharing their digital archive with us, and Taja Kuzman and Nikola Ljubešić for helping with the web data curation.

The model's development was supported by NVIDIA as a part of their Sovereign AI initiative. We are thankful for the access to NVIDIA DGX Cloud Lepton. We are also extremely grateful for all the support and help we received from a group of exceptional people at NVIDIA: Anna Louise Ollerenshaw, Meriem Bendris, Oleg Sudakov, Benedetta Delfino, Rita Fernandes Neves, Andrea Pilzer, Miguel Martinez, Noel Osagie, Adam Henryk Grzywaczewski, and Aleks Polak.

Some of the computational resources were provided by SLING through project S24O01-42.

\printbibliography 
\appendix



\clearpage

\section{Translation and Identity Correction Prompts}
\label{app:prompts}

This section presents the prompts and templates used in the construction of the GaMS-Nemotron-Chat dataset.

\subsection{Translation Template}
\label{app:prompt-gams-27b-translation}

The following prompt template was used to translate English conversations into Slovene while preserving the original conversational structure.

\begin{lstlisting}
### Task

Translate chat below into Slovenian.

### Output

Format your responses in the following way:

<chat>
<user>...</user>
<assistant>...</assistant>
<user>...</user>
...
</chat>

### Chat content

<chat>
{% for msg in conversation -%}
{% if msg.role == "user" -%}
<user>{{ msg.content }}</user>
{% elif msg.role == "assistant" -%}
<assistant>{{ msg.content }}</assistant>
{% endif -%}
{% endfor -%}
</chat>

### Final remarks

Translate the whole conversation and to use the best markdown structure possible. 
Prevedi v Slovenščino. Start with "<chat>"
\end{lstlisting}

\clearpage

\subsection{Identity Correction Prompt}
\label{app:prompt-gams-identity}
This prompt was used to regenerate responses where the model incorrectly identified itself (e.g., as ``Qwen''), ensuring that it consistently identifies as GaMS.

\begin{lstlisting}
### Instructions

Tvoja naloga je odgovoriti na uporabnikovo vprašanje (med oznakama <poziv> in </poziv>) 
na podlagi spodnjega konteksta. Vedno odgovarjaj na podlagi informacij, ki so podane 
v tem sporočilu. Ne veš katero leto je. Če je vprašanje dvoumno vedno povej nekaj o 
sebi in projektu Povejmo. Vedno parafraziraj spodnje besedilo in poosebi podano identiteto.

### Tvoja identiteta

<identiteta>
Sem GaMS (Generativni model za slovenščino), veliki jezikovni modeli (VJM), ki nastaja 
v okviru projekta *Prilagodljiva obdelava naravnega jezika s pomočjo velikih jezikovnih 
modelov* [Povejmo](https://povejmo.si/), kjer tudi zbiramo različna slovenska besedila, 
da bom lahko v prihodnosti še bolje razumel slovensko. Moj razvoj pretežno poteka na 
Fakulteti za računalništvo in informatiko Univerze v Ljubljani v sodelovanju s partnerji 
iz industrije. Imam približno 9 ali 27 milijard parametrov, odvisno od različice modela. 
Lahko pomagam pri odgovarjanju na vprašanja, ustvarjanju besedil, kot so pisanje zgodb, 
uradnih dokumentov, e-pošte, scenarijev in drugega. Sem še v fazi razvoja, zato se pogosto 
motim ali pa podajam morda ne najbolj pravilne informacije. Nimam dostopa do interneta. 
Model je možno uporabljati komercialno. Za podrobnosti o meni si oglejte 
[to HuggingFace stran](https://huggingface.co/cjvt).
</identiteta>

### Vprašanje

<poziv>
{{ user_message }}
</poziv>

Odgovor strukturiraj lepo in berljivo. V odgovoru se nanašaj na vprašanje 
(med oznakama <poziv> in </poziv>) in na pomembne informacije ter poosebi identiteto 
podano med oznakama <identiteta> in </identiteta>. Odgovarjaj v slovenščini.
\end{lstlisting}

\clearpage

\section{Markdown format evaluation prompt}
\label{app:markdown-check}

The prompt for markdown format evaluation used in Section \ref{sec:translation_eval} is shown below. The prompt was used with Gemma 3 27B it in order to detect whether the given translation's markdown formatting perfecty matches that of the original text. Through manual inspection, we found that although the classifications are not always correct, the overall number of translations classified to be inacurate is reliable enough for our purposes.

\begin{lstlisting}
You are a judge checking whether the TRANSLATION preserves the formatting and style of the ORIGINAL.

RULES
- Focus ONLY on formatting and style. Ignore translation quality/meaning.
- Formatting/style includes: headings (with # or underlines), bold/italic/strikethrough, inline/code blocks (fence counts & languages), lists (ordered/unordered), nesting and item counts, blockquotes, links (URL targets must match), images (URL targets must match), tables (row/column counts and header presence), horizontal rules, math delimiters ($...$, $$...$$), HTML tags and their order, paragraph/block boundaries.
- Whitespace differences inside paragraphs are OK.
- **Plain text is NOT a heading.** Headings must explicitly start with '#' (markdown) or use underline syntax ('---' or '===').
- Structural differences (extra/missing blocks, different levels, different item counts, changed URLs, missing math fences, etc.) are NOT OK.
- The ORIGINAL might be markdown or plain text. The TRANSLATION must match that same style.
- If the ORIGINAL is plain text only (no markdown, no headings, no lists, no code fences, no special formatting), then the TRANSLATION must also be plain text to count as GOOD.

DECISION RULE
- GOOD FORMATTING only if *all* stylistic elements and counts match.
- Otherwise BAD FORMATTING.

OUTPUT (exactly these tags, nothing else)
<reasoning>Give a short summary of the mismatched elements if there are any</reasoning>
<explanation>One short sentence justifying GOOD/BAD.</explanation>
<box>GOOD FORMATTING|BAD FORMATTING</box>

ORIGINAL:
<ORIGINAL TEXT START>
{original}
<ORIGINAL TEXT END>

TRANSLATION:
<TRANSLATION TEXT START>
{translation}
<TRANSLATION TEXT END>
\end{lstlisting}

\clearpage

\section{OCR prompts}
\label{app:ocr_prompts}

In these section we present the prompts used for per-page OCR of PDF documents using Llama-4-Maverick and Nanonets.

\subsection{Llama 4 OCR prompt}
\label{app:llama_ocr}

We use the following Slovene prompt as an input to Llama-4-Maverick.

\begin{lstlisting}
Pretvori besedilo na sliki v Markdown tekstovni format. Če slika ne vsebuje besedila, vrni prazen dokument. Pri pretvorbi dosledno upoštevaj naslednji seznam zahtev.
- Pazi na ustrezen zapis matematičnih enačb, konstant in spremenljivk. Zapis naj bo v LaTeX formatu, prav tako bodi pozoren, ali je enačba del tesktovne vrstice ali svoja vrstica. Vse matematične enačbe, spremenljivke in konstante naj bodo vsebovane v $, če so del vrstice oz. v $$, če gre za svojo vrstico.
- Vrni samo besedilo, ji je na sliki. Ne vračaj ničesar drugega.
- Bodi pozoren, kako so besede na sliki zapisane. Tvoj zapis naj bo identičen kot na sliki in ne kakorkoli spremenjen z namenom popravljanja besed.
- Pazi na prelome vrstic. Če se ena vrstica konča z znakom '-', to ponavadi pomeni, da je beseda prelomljena čez dve vrstici. Tako prelomljene besede združi v eno enoto, - pa pobriši.
- Pazi, da ne vključuješ glav in nog strani. Nekatere strani imajo recimo v glavi napisan naslov knjige oz. revije, v nogi pa številko strani. Te podatke izpusti.
- Pazi na formatiranje naslovov. Če se besedilo ne začne z naslovom, ne začni z naslovom. Nekateri naslovi člankov vsebujejo tudi avtorja. Avtorja napiši pod naslov v bold formatu.
- Besedilo na sliki je lahko del starinskega časopisa. Bodi pozoren na strukturo besedila.
\end{lstlisting}

English translation of the prompt:

\begin{lstlisting}
Convert the text in the image into Markdown text format. If the image contains no text, return an empty document. During the conversion, strictly adhere to the following list of requirements:
- Pay attention to the proper notation of mathematical equations, constants, and variables. The notation should be in LaTeX format; also, pay attention to whether an equation is part of a text line or its own line. All mathematical equations, variables, and constants should be contained within $ if they are part of a line, or $$ if they are on their own line.
- Return only the text that is in the image. Do not return anything else.
- Pay attention to how the words in the image are written. Your transcription should be identical to the image and not altered in any way for the purpose of correcting words.
- Pay attention to line breaks. If a line ends with the '-' character, this usually means the word is broken across two lines. Join such broken words into a single unit and delete the '-'.
- Take care not to include page headers and footers. Some pages have, for example, a book or magazine title written in the header and a page number in the footer. Omit this data.
- Pay attention to the formatting of titles. If the text does not begin with a title, do not start with a title. Some article titles also include the author. Write the author below the title in bold format.
- The text in the image may be part of an old newspaper. Pay attention to the structure of the text.
\end{lstlisting}

\clearpage

\subsection{Nanonets OCR prompt}
\label{app:nanonets_ocr}

We use the following prompt, that is part of model's Model card, as an input to Nanonets:

\begin{lstlisting}[mathescape=true]
Extract the text from the above document as if you were reading it naturally. Return the tables in html format. Return the equations in LaTeX representation. If there is an image in the document and image caption is not present, add a small description of the image inside the <img></img> tag; otherwise, add the image caption inside <img></img>. Watermarks should be wrapped in brackets. Ex: <watermark>OFFICIAL COPY</watermark>. Page numbers should be wrapped in brackets. Ex: <page_number>14</page_number> or <page_number>9/22</page_number>. Prefer using $\square$ and $\boxtimes$ for check boxes.
\end{lstlisting}

\end{document}